\documentclass[twocolumn, switch]{article} 

\usepackage{preprint}

\usepackage{amsmath, amsthm, amssymb, amsfonts}

\usepackage[numbers,square]{natbib}
\usepackage[utf8]{inputenc}	
\usepackage[T1]{fontenc}	
\usepackage{xcolor}		
\usepackage[colorlinks = true,
            linkcolor = purple,
            urlcolor  = blue,
            citecolor = cyan,
            anchorcolor = black]{hyperref}	
\usepackage{booktabs} 		
\usepackage{nicefrac}		
\usepackage{microtype}		
\usepackage{lineno}		
\usepackage{float}			
\usepackage{array}
\usepackage{wrapfig}
\usepackage[font=small,labelfont=bf,skip=0pt]{caption}
\DeclareCaptionLabelFormat{andtable}{#1~#2  \&  \tablename~\thetable}
\usepackage{amsmath}
\usepackage{subcaption}
\usepackage{graphicx}

\newenvironment{conditions}
  {\par\vspace{\abovedisplayskip}\noindent\begin{tabular}{>{$}l<{$} @{${}={}$} l}}
  {\end{tabular}\par\vspace{\belowdisplayskip}}

\usepackage{lipsum}		

\usepackage{newfloat}
\DeclareFloatingEnvironment[name={Supplementary Figure}]{suppfigure}
\usepackage{sidecap}
\sidecaptionvpos{figure}{c}

\usepackage{titlesec}
\titlespacing\section{0pt}{12pt plus 3pt minus 3pt}{1pt plus 1pt minus 1pt}
\titlespacing\subsection{0pt}{10pt plus 3pt minus 3pt}{1pt plus 1pt minus 1pt}
\titlespacing\subsubsection{0pt}{8pt plus 3pt minus 3pt}{1pt plus 1pt minus 1pt}

\usepackage{tikz,xcolor,hyperref}

\definecolor{lime}{HTML}{A6CE39}
\DeclareRobustCommand{\orcidicon}{
	\begin{tikzpicture}
	\draw[lime, fill=lime] (0,0) 
	circle [radius=0.16] 
	node[white] {{\fontfamily{qag}\selectfont \tiny ID}};
	\draw[white, fill=white] (-0.0625,0.095) 
	circle [radius=0.007];
	\end{tikzpicture}
	\hspace{-2mm}
}
\foreach \x in {A, ..., Z}{\expandafter\xdef\csname orcid\x\endcsname{\noexpand\href{https://orcid.org/\csname orcidauthor\x\endcsname}
			{\noexpand\orcidicon}}
}

\title{Dataset: Impact events for Structural Health Monitoring of a thin plate}

\usepackage{authblk}

\author[1]{Ioannis Katsidimas\orcidA{}}
\author[2]{Thanasis Kotzakolios}
\author[1,2]{Sotiris Nikoletseas}
\author[1]{Stefanos H. Panagiotou\orcidB{}}
\author[1]{Konstantinos Timpilis}
\author[1]{Constantinos Tsakonas}
\affil[1]{Computer Engineering and Informatics Department, University of Patras, Greece \authorcr Email: {\tt \{ikatsidima, spanagiotou, timpilis, st1059666\}@ceid.upatras.gr}\vspace{0.5ex}, 
}
\affil[2]{Mechanical Engineering and Aeronautics Department, University of Patras, Greece \authorcr Email: {\tt  \{kotzakol\}@upatras.gr} \vspace{0.5ex}}
\affil[3]{Computer Technology Institute and Press “Diophantus” (CTI), Greece \authorcr Email: {\tt \{nikole\}@cti.gr}\vspace{0.5ex}}

\begin{document}

\twocolumn[ 
  \begin{@twocolumnfalse} 
  
\maketitle

\begin{abstract}

Nowadays, more and more datasets are published towards research and development of systems and models, enabling direct comparisons, continuous improvement of solutions, and researchers engagement with experimental, real life data. However, especially in the Structural Health Monitoring (SHM) domain, there are plenty of cases where new research projects have a unique combination of structure design and implementation, sensor selection and technological enablers that does not fit with the configuration of relevant individual studies in the literature. Thus, we share the data from our case study to the research community as we did not find any relevant repository available. More specifically, in this paper, we present a novel time-series dataset for impact detection and localization on a plastic thin-plate, towards Structural Health Monitoring applications, using ceramic piezoelectric transducers (PZTs) connected to an Internet of Things (IoT) device. The dataset was collected from an experimental procedure of low-velocity, low-energy impact events that includes at least 3 repetitions for each unique experiment, while the input measurements come from 4 PZT sensors placed at the corners of the plate. For each repetition and sensor, 5000 values are stored with 100 KHz sampling rate. The system is excited with a steel ball, and the height from which it is released varies from 10 cm to 20 cm. The dataset is available in GitHub (https://github.com/Smart-Objects/Impact-Events-Dataset).
\end{abstract}
\vspace{0.35cm}

  \end{@twocolumnfalse} 
] 


\section{Introduction}
During the last years, the use of IoT technology has proven to be inevitable for the critical services it can offer in structural monitoring and operation, due to an abundance of benefits it provides such as low cost, scalability, interoperability with other systems and lately the AI extensive utilization.

Many times we come across with systems that claim to be categorized as IoT, costing some thousands of dollars, and are directly compared to previous colossal and costly approaches. In our opinion, IoT must inherently maintain in its nature cost effectiveness, in order to support a larger covering area and more data sources. Primarily, IoT advantages relies on the amount of data we are able to collect from points of operational interest, rather than the quality of each particular measurement. Thus, complementary technologies that complete IoT's gaps have emerged such as ML on IoT data, outliers detection, calibration and measurement accuracy improvement. 

IoT has also given a great boost at the ``Structural Health Monitoring" domain, which usually refers to the process of implementing a damage detection strategy for aerospace, civil or mechanical engineering infrastructures, while it is not that widespread for simple and smaller structural elements as parts of bigger systems. However, as the cost remains low this is feasible even for the individual building elements that together can form a bigger distributed and interconnected monitoring system.

Most SHM algorithms and datasets most often involve information on both damaged and undamaged state information.This requires extensive data to be readily available from the sensing systems, the physics-based models, or the experiments. Nevertheless, this is not possible in many cases, and the current information for damage state is limited, if not unavailable. This work focuses on the external impact events that often occur in engineering structures such as aircraft and wind turbine blades. Therefore, the fast and accurate detection of impact forces and the localization of impact location can be used before the damage assessment of the structure.


\vspace{0.1cm}
\textbf{Limitations of existing datasets.}
Despite the large number of research works on SHM, the vast majority do not share their datasets. Azimi et al. \cite{azimi2020data}, provide an extensive list of those publications that do share their datasets (with vibration and mostly vision-based data) that have been recently used in deep learning-based SHM. However, for impact detection and localization in plate structures with PZT sensors, which is also a very well studied problem in the literature (for example \cite{act10050101, 8653342,doi:10.1177/14759217221098569,DIPIETRANGELO2023109621, de2018new}), there are no openly available datasets. Regarding some indicative open access datasets (for any SHM-related problem),  Bechhoefer et al. \cite{bechloefer} have published data for Wind Turbine High-Speed Bearing Prognosis and also Figueiredo et al. \cite{osti961604}   from Los Alamos National Lab have published standardized datasets intending to familiarise users with feature extraction and statistical modelling for feature classification in the context of SHM. Teloli et al. \cite{teloli2019new} have a published dataset named as UNESP-BERT for bolted joint SHM based on vibration tests. Finally, Marzani et al. \cite{marzani2020open}, have made available a dataset for benchmarking guided waves on a composite full-scale outer section of an aircraft wing. However, utilizing the aforementioned datasets the algorithms that can be trained focus on to detect and sometimes, but not always, locate damage while they do not consider environmental and operational factors and only rely on specific damages (change in geometry, material) that occur in the structure.

\vspace{0.1cm}
\textbf{Our contributions.}
To the best of our knowledge, we are the first, to publish a public dataset that contains PZT sensors measurements concerning low-velocity, low-energy impact events in a thin plastic plate. In addition, we also contribute with our methodology on data collection using an SHM IoT system with resource constraints (based on Arduino NANO 33 MCU), as opposed to the majority of the literature that uses Oscilloscopes for data acquisition. This concept of an MCU-based system for data collection in SHM is especially important nowadays, due to the fast rise of extreme-edge and embedded machine learning practices solutions that enable a variety of real-time data-driven SHM applications. Finally, we wish to highlight that by using this specific Microcontroller Unit (MCU) and sensors, the proposed implementation aims for an overall low-cost data collection solution.

\vspace{0.1cm}
\textbf{Potential Dataset Use Cases.}
The Impact Events dataset contains measurements from four PZT sensors bonded in a thin plastic (acrylic) structure after dropping a steel ball from different heights. This dataset may be therefore utilized from the community to conduct research in the following indicative use cases:
\begin{itemize}
    \item \textbf{Impact detection}. The aim is to leverage the sensor measurements for distinguishing if a significant impact event has been occurred or if the measurements have been excited from irrelevant causes.
    \item \textbf{Impact localization}. Locate the area of the occurred impact by two main approaches. The first one is to define a grid in the surface of the plate that consists of arbitrary distinct tiles and the second one is to locate the exact X- and Y-coordinates  of the steel ball impact. These two approaches can then formulate and be solved as an example via supervised machine learning classification and regression algorithms respectively.
    \item \textbf{Steel ball height classification}. Since the steel ball experiments have been conducted from several heights, a potential use case is to determine the height from which the steel ball was released, in a new unseen experiment.
    \item\textbf{Force of impact estimation}. The goal is to classify the energy/force levels of impact.  Knowledge of the impact energy gives an understanding of the severity of the impact; a severe impact is associated with composite damage, such as delamination and fibre cracking.
\end{itemize}


\section{Methodology}
In this section we provide the methodology that was followed to create the Impact Events dataset, in the context of statistical pattern recognition of impact detection and quantification. First, we describe the design of the thin plate structure. Afterwards, we provide the details for the design of experiments, how and why they were performed and then we outline the procedure of the data acquisition with the edge sensing device (Arduino) along with the overall challenges faced to perform the data collection.



\subsection{Thin Plate and Sensors Setup}

\begin{wrapfigure}[11]{r}{0.25\textwidth}
    \includegraphics[width=\linewidth]{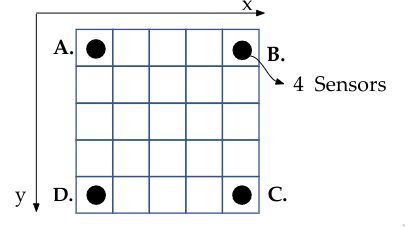}
    \caption{Thin plate model with the sensors and the impact localization areas}
    \label{fig:plate}
\end{wrapfigure}

The structure that is used is a thin acrylic plate $P$ with dimensions of 300mm x 300mm and 4mm thickness and with four Piezoelectric transducers (PZTs) bonded at the corners, as shown in Fig. \ref{fig:plate}. The grid is  separated in a grid of 25 identical 60mm x 60mm square tiles, to allow the modeling of impact localization problems with classification approaches. However, since the grid tiles are essentially derived from the X and Y-coordinates of the impact, the number and size of tiles can be arbitrarily formulated to fewer or more labels depending on the research purposes. The set of the squares is denoted as $Q=\{q_{1,1},\ q_{1,2},\ ...,\ q_{1,5},\ q_{2,1},\ q_{2,2},\ ...,\ q_{2,5},\  q_{3,1},\ ...,\ q_{5,4},\ q_{5,5}\}$ where x and y at $q_{x,y}$ is the row and column square respectively, thus PZT A is deployed at $q_{1,1}$, PZT B at $q_{1,5}$, PZT C at $q_{5,5}$, and PZT D at $q_{5,1}$ (and they are denoted as sensor[A|B|C|D] in the following sections).

\begin{figure}[!htbp]
    \centering
    \includegraphics[height=5cm, width=1.0\columnwidth]{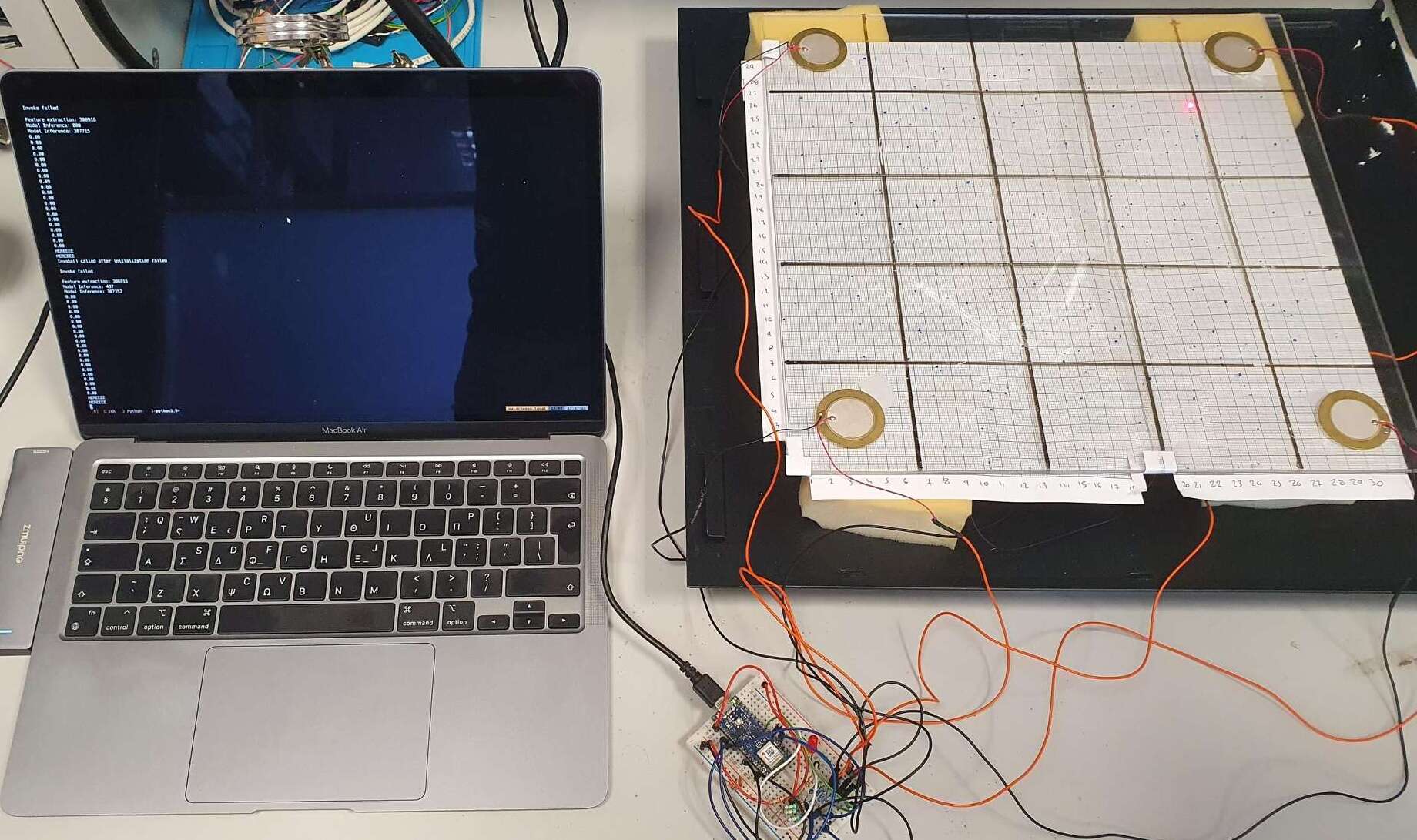}
    \caption{Image of the experimental plate setup. The laptop is used to store the data that the arduino device reads from the piezoelectric sensors.}
    \label{fig:Visual-plate}
\end{figure}

\begin{figure}[!h]
    \centering
   \includegraphics[scale=0.2]{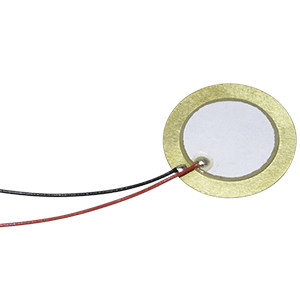}
    \qquad
     \begin{tabular}[b]{cc}\hline
          Parameter & Units \\ \hline
          operating voltage & max. 30V  \\
          resonant frequency & typ. 2,6 Hz \\
          weight & max. 2.0 g \\
          dimensions & Ø35 x 0.53 mm \\
    \end{tabular}
    \captionlistentry[table]{A table beside a figure}
    \captionsetup{labelformat=andtable}
    \caption{Ceramic piezoelectric transducer CEB-35D26 and its main specifications.}
    \label{table:pzt}
  \end{figure}

In Figure \ref{fig:Visual-plate} we provide an image of the experimental plate setup, while in Table \ref{table:pzt} we provide the attributes of the Ceramic piezoelectric transducer CEB-35D26 \cite{pzt} that was used.
\\


Below we provide the formula for correlating the force applied to the output voltage of the piezoelectric sensor.

 \begin{equation}
    F=V*d_{33}*C
 \end{equation}
 where
 \begin{conditions}
 F      &    Applied force\\
 V      &    Output voltage\\
 d_{33} &    Piezoelectric constant\\
 C      &    Stiffness factor of ceramic\\
 \end{conditions}
     
For the CEB-35D26, $d_{33}=\ 460\ pC/N$ and $C=\ 60*10^9\ N/m^2$

\subsection{Design and execution of experiments}
The data acquisition portion of the SHM process involves selecting the excitation methods; the sensor types, numbers, and locations; and the data acquisition/storage/ processing/transmittal hardware.

\begin{figure}[h]
    \centering    \includegraphics[scale=0.3]{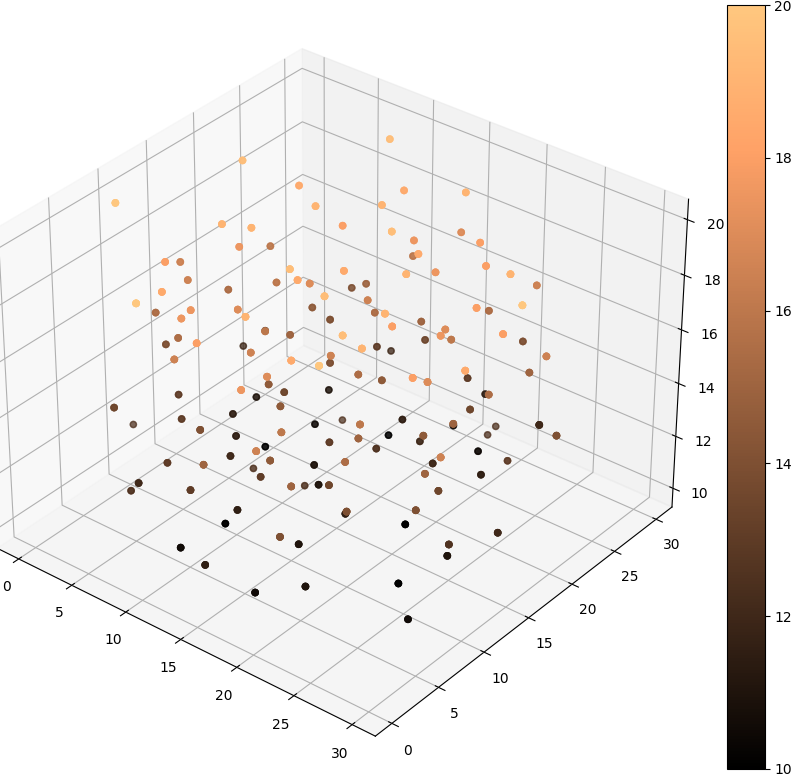}
    \caption{3D representation of impact coordinates and fall height combinations, based on Sobol sequences}
    \label{fig:sobol}
\end{figure}

As a first step, a design of experiments was performed to associate the sensors response to the parameters defining the ball impact. The experiment design or Design of Experiments (DOE) is a process of generating test sequences. The test sequences consist of operating points of input variables of a system. The experiment is designed to maximize the coverage of test sequences to cover the entire operating region of a system. The design is maximized in such a way that the model will be able to replicate all behavior of the system. Namely, the x-coordinate, the y-coordinate and the fall height. DOEs are categorized in space-filling and optimal DOEs. This particular problem was treated as an unknown system. Therefore, a space-filling DOE was selected, and more specifically the Sobol sequence. Sobol sequences or also known as LPT sequences is categorized as a space-filling design. This design is a quasi-random sequence, in which the test sequences generated are randomly planned in the design boundary. The test sequences are iteratively generated in a uniform distribution. The most fascinating characteristic of Sobol sequences is that it generates test sequences in high degree of scatter while avoiding overlapping of previous test sequences. The deterministic quasi- random characteristic of Sobol sequences enables progressive augmentation of the test sequences \cite{10.1145/42288.214372,867640}. In Figure \ref{fig:sobol}, we provide 3D representation of impact coordinates and fall height combinations based on Sobol sequences.

\begin{figure}[h]
    \centering    
    \includegraphics[scale=0.3]{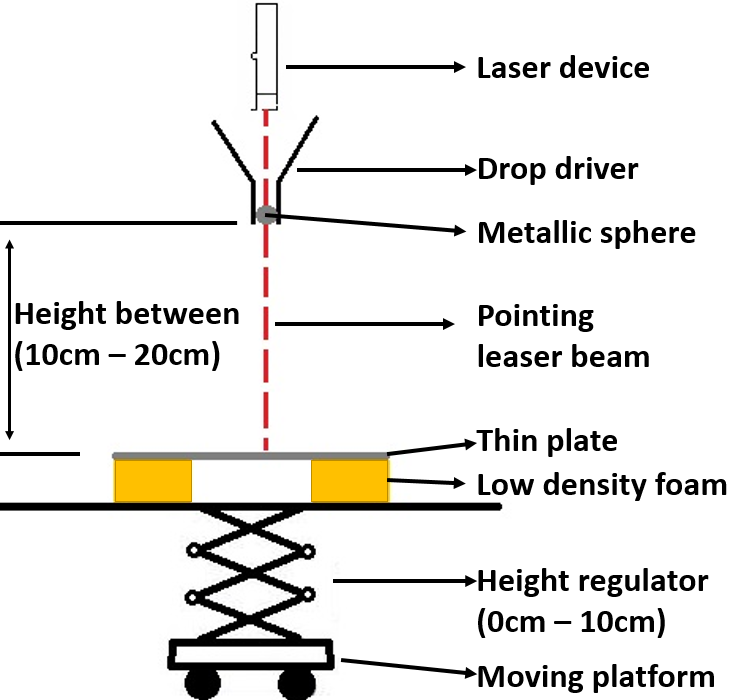}
    \caption{Visualization of the steel ball drop setup.}
    \label{fig:stell-ball-drop}
\end{figure}

In Figure \ref{fig:stell-ball-drop}, we depict the setup for dropping the steel ball. In specific, the impact event is produced by a steel ball $B$ (9.5 mm diameter, 3.53 gr weight), which is released from a height distance that varies from 10 cm to 20 cm, with 0.5 cm interval, from the $P$, and performs a free-fall (the initial state of the sphere has zero acceleration with the help of the Drop driver). The Laser device and the Drop driver are statically mounted and the change of impact location takes places by moving the plate in the desired X- and Y-coordinates, which are verified using the Laser beam. We also note that in the experimental procedure we do not perform impact events in the square tiles where the sensors are included (i.e., in $q_{1,1}$, $q_{1,5}$, $q_{5,1}$, $q_{5,5}$), as the latter return extreme and noisy values. Overall, each experiment is repeated at least three times in order to enhance the dataset and ensure the robustness of the experiment, resulting in total 771 experiments (multiple repetitions for 200 distinct impacts).

\subsection{Data collection with edge sensing device}
\begin{figure}[h]
    \centering    
    \includegraphics[scale=0.08]{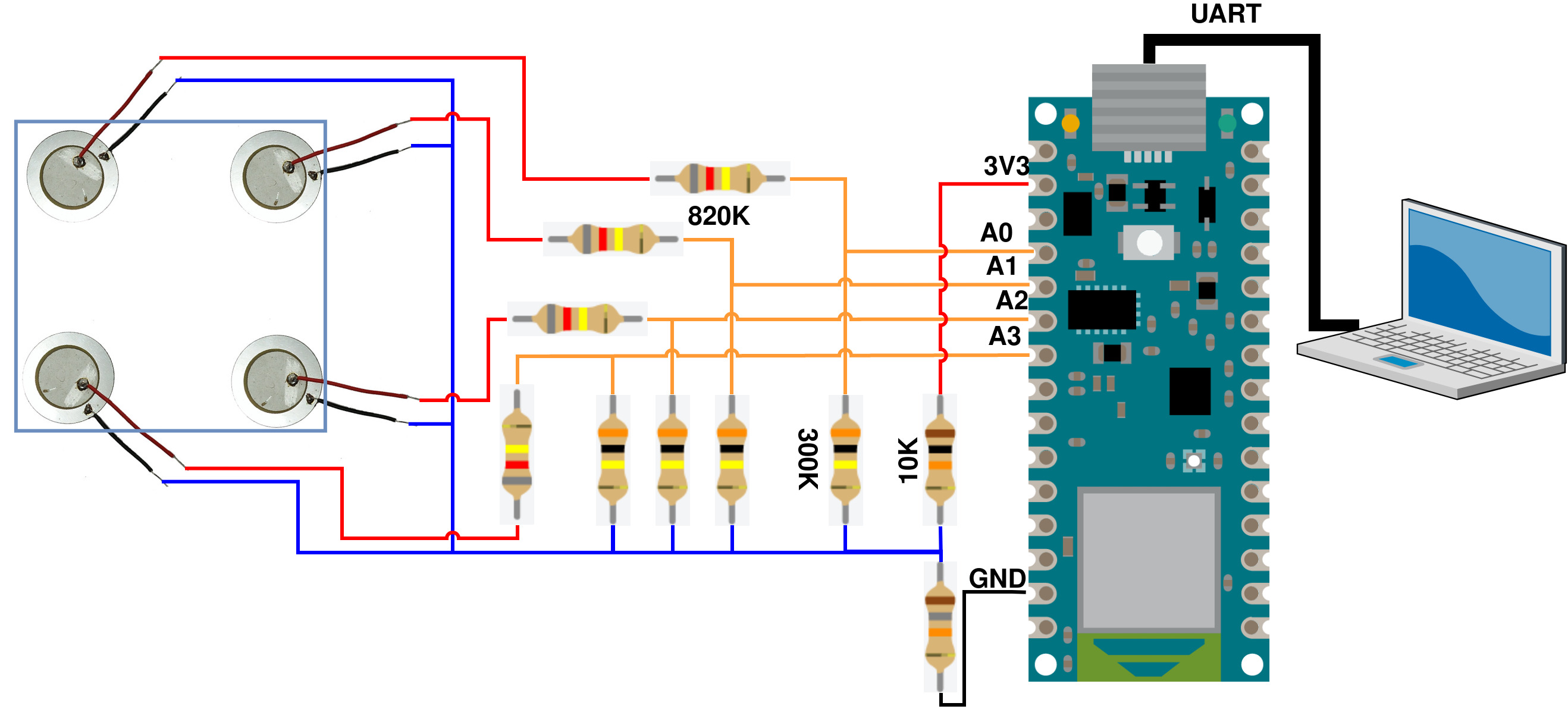}
    \caption{PZT modules connected to Arduino NANO 33 BLE.}
    \label{fig:arduino}
\end{figure}

The data is collected using the microcontroller, which sends over the UART bus the sensed values to store them permanently in an offline workstation (e.g. laptop). Figure \ref{fig:arduino} depicts the schematic of the plate-sensor-arduino-laptop setup. The sensors are connected to the analog inputs of the device and the sampling frequency of the Analog-to-Digital Converter (ADC) is set at 100KHz, to ensure that the impact phenomenon is captured as detailed as possible. Additionally, the signal that we capture is a shock that stimulates the eigenfrequencies of the structure for a small period of time, thus the correct way to treat the captured signal for a specific experiment is to consider it as one instance of 20k samples (4 sensors x 5k samples each).

\subsection{Challenges}

The collection of the proposed dataset contains a number of challenges, namely: Data quality (addressed by executing multiple repetitions of the same unique experiments and defining the optimal impact samples using the Sobol algorithm, Reliable impact stimulation methodology (addressed by our steel ball drop setup in \ref{fig:stell-ball-drop} to ensure consistent labels and reproducible experiments), PZT sensor stability (addressed by manual testing of each PZT sensor's responses before including them in the final experiments) and  high frequency data sampling with automated acquisition and processing of samples. Below, we elaborate on how we addressed the last challenge.
\begin{wrapfigure}[17]{r}{0.21\textwidth}
    \centering 
    \centerline{\includegraphics[width=0.55\linewidth]{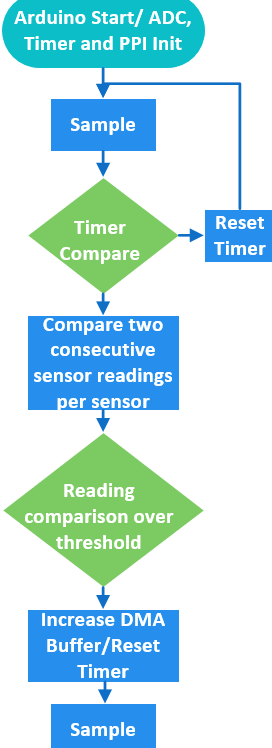}}
    \caption{Sampling Procedure}
    \label{fig:sampling_flow_chart}
\end{wrapfigure}

The whole process of  the sensors reading is implemented on the hardware level, exploiting the Arduino's Successive Approximation ADC (SAADC), and thus the processor is not responsible for enabling the analog channels and storing the values in the memory repeatedly. Although SAADC's sampling rate is over the mark of 15KHz and operates with multi-channel mode, it is important to note that the Arduino introduces memory writing abnormalities, due to synchronization problems updating SAADC's storing buffer index and size. Hence, the main solution to overcome this problem is to transfer the data to a second data structure as soon as reading of all the available channels is finished and then restart the SAADC to reinitialize the buffer. However, we exploit the Programmable Peripheral Interface (PPI), interconnecting the SAADC and a Timer to synchronize the needed operations to update the size of the SAADC's buffer using timer triggers. By using the PPI, the SAADC's restarting is avoided. The SAADC's internal resistance is disabled and the gain is set to $\frac{1}{6}$. The reference voltage is 0.6 Volt.

For the signal recording, an auto-trigger mechanism is developed on the device to initiate the acquisition procedure as shown in Figure \ref{fig:sampling_flow_chart}. Everything regarding the sampling procedure is implemented on the hardware level for efficiency. The only action that the processor performs at this stage is to check if the relative change in two successive readings is greater than a threshold set to 10\%.

\section{Exploratory data analysis}

\subsection{Outline of the dataset}
In the published data repository \cite{impactdataset}, the directory of the Impact Events dataset is organized as follows. \emph{DoE\_Data} folder contains one folder for each different height of the impact ball fall (ranging from 10.0 to 20.0 cm with an interval of 0.5 cm). Then each of those folders contains one .csv file per experiment, based on the generated Sobol sequences (\emph{sobol\_[Height]\_[Sobol\_ID].csv}). \emph{Feature\_extraction} folder contains the \emph{features.json} file that includes the index of the statistical features to be extracted (see the following section) and \emph{feature\_augmented\_dataset.csv} that contains the extracted features (as columns) and the rest of impact information (x,y, height, position) per experiment (rows). Finally, the \emph{merged\_dataset.csv} file contains the unified dataset from all the sensor measurements across all the experiments, and its structure is represented in Table \ref{table:DatasetOverview}.

\begin{figure}[ht]
  \begin{subfigure}{0.45\columnwidth}
  \includegraphics[width=\textwidth]{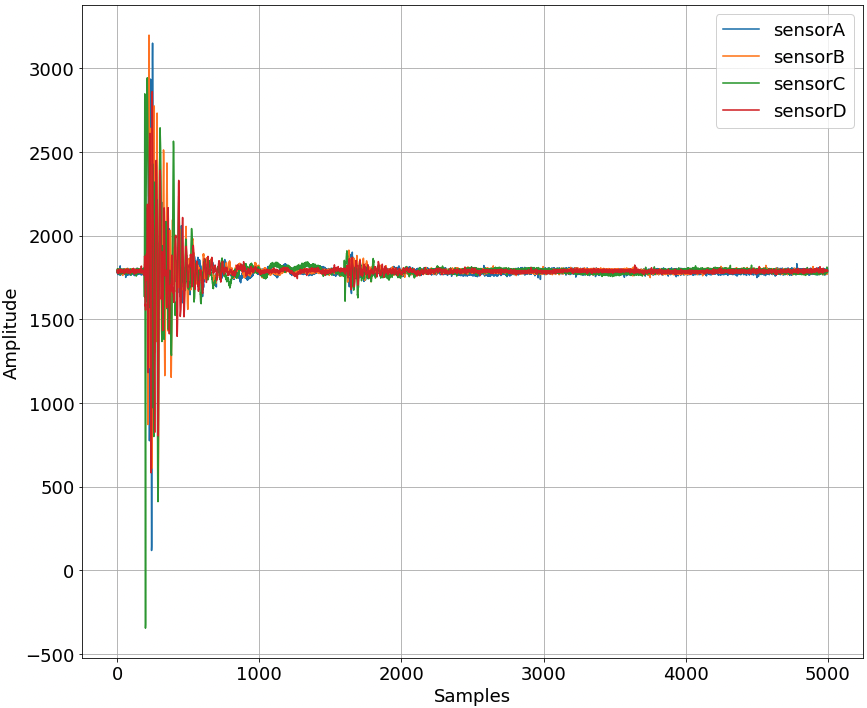}
  \caption{Raw data over 5k samples}
  \end{subfigure}
  \hfill
  \begin{subfigure}{0.45\columnwidth}
  \includegraphics[width=1.03\textwidth]{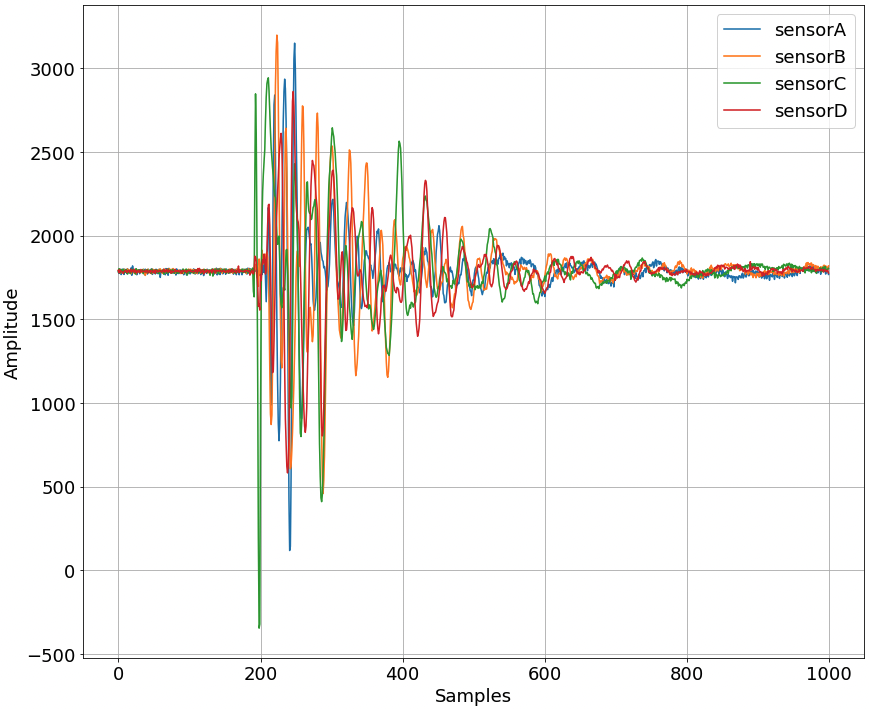}
  \caption{Raw data over 1k samples} 
  \end{subfigure} 
  \begin{subfigure}{0.45\columnwidth} 
  \includegraphics[width=\textwidth]{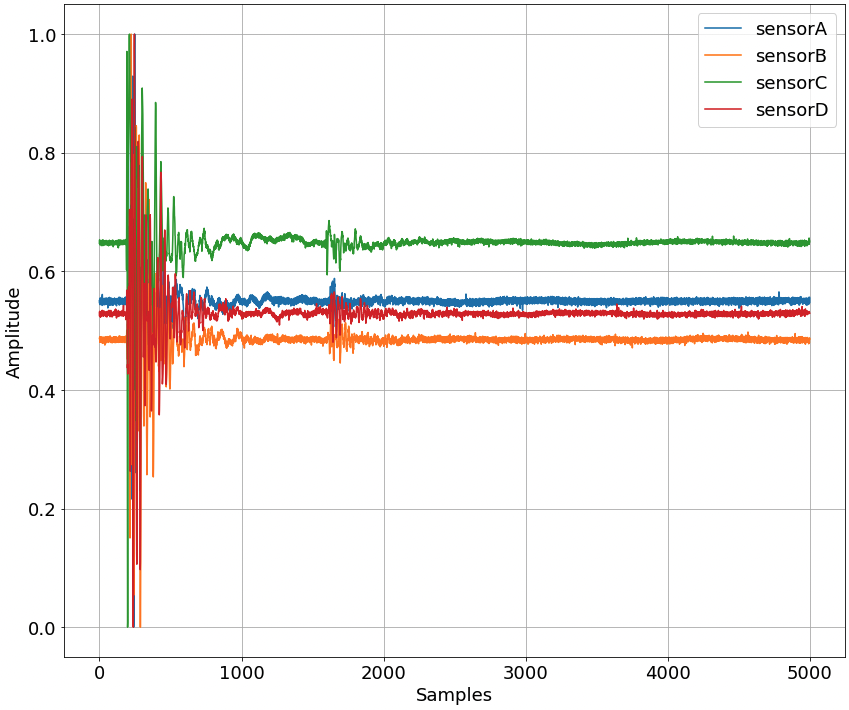}
  \caption{Min-max scaling of data over 5k samples} 
  \end{subfigure}  
  \hfill 
  \begin{subfigure}{0.45\columnwidth} 
  \includegraphics[width=1.03\textwidth]{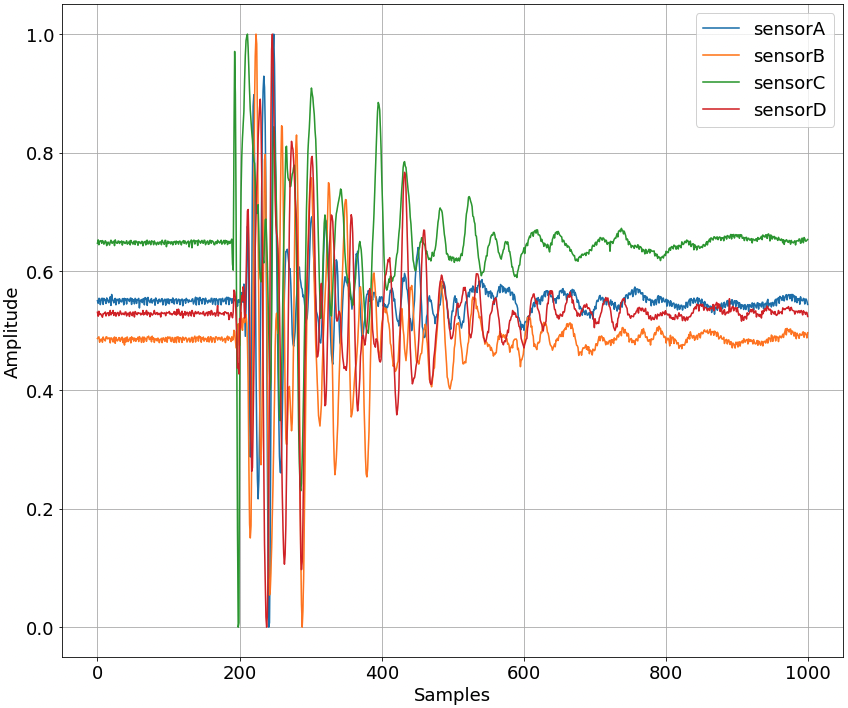}
  \caption{Min-max scaling of data over 1k samples}
  \end{subfigure}
  \caption{Visualizing measurements from all sensors, over raw and min-max-scaling values and all 5k or 1k samples}
  \label{figure:sensor-values}
  \end{figure}

Furthermore, in Table \ref{table:statistics} we provide common statistical  measures to describe the PZT sensor measurements, while in Figure \ref{figure:sensor-values} we present examples of the sensor values, with or without min-max scaling and for the first 1k samples or all the 5k samples.

\begin{table}[h]
\centering
\begin{tabular}{l|llll}
Statistic & sensorA & sensorB & sensorC & sensorD \\ 
\hline
count & 3855000 & 3855000 & 3855000 & 3855000 \\
mean & 1790.466 & 1794.333 & 1793.815 & 1792.419 \\
std & 86.822 & 101.574 & 83.672 & 100.586 \\
min & -357 & -385 & -352 & -371 \\
25\% & 1775 & 1781 & 1780 & 1782 \\
50\% & 1796 & 1799 & 1799 & 1797 \\
75\% & 1810 & 1812 & 1812 & 1809 \\
max & 4048 & 4057 & 4040 & 4055
\end{tabular}
\caption{Descriptive statistics for the four PZT sensor measurements}
\label{table:statistics}
\end{table}

\begin{table}[h!]
\resizebox{\columnwidth}{!}{%
\begin{tabular}{|l|l|l|l|l|l|l|l|l|l|l|l|}
\hline
SensorA & SensorB & SensorC & SensorD & sampleNo & typeofimpact & x & y & height & position & ID \\ \hline
\end{tabular}%
}
\caption{Dataset format}
\label{table:DatasetOverview}
    \vspace{-1.0em}
\end{table}



\subsection{Feature Extraction and Importance}
To further analyze the dataset we extract several statistical measures to provide additional meaningful insights and representation of the data. For example, one potential use of the extracted features is to used them as a tabular input in machine learning models to localize the area of impact i the palte.  The features are calculated in the offline workstation (e.g. laptop) using the tsfresh library \cite{christ2018time}, applied on the \emph{merged\_dataset.csv}. A detailed list of the features can be found in the data repository.

More specifically, the features are calculated per experiment and each statistical measure is calculated for each PZT sensor, resulting in  \emph{feature\_augmented\_dataset.csv} file that contains 128 features (4 PZT sensors x 32 statistical measures). Overall, this tabular dataset contains 771 rows corresponding to the unique impact experiments, 128 columns corresponding to each feature and the target columns, X and Y for impact coordinates, Height for the ball fall height and position for the impact area $q_{x,y}$.
Usually only a subset of features is important and representative of the dataset to result in high accuracy for machine learning models. Therefore, after extracting several statistical features, we present the most important ones, based on the ANOVA statistical method.

\begin{figure}[h]
    \centering    \includegraphics[width=\columnwidth]{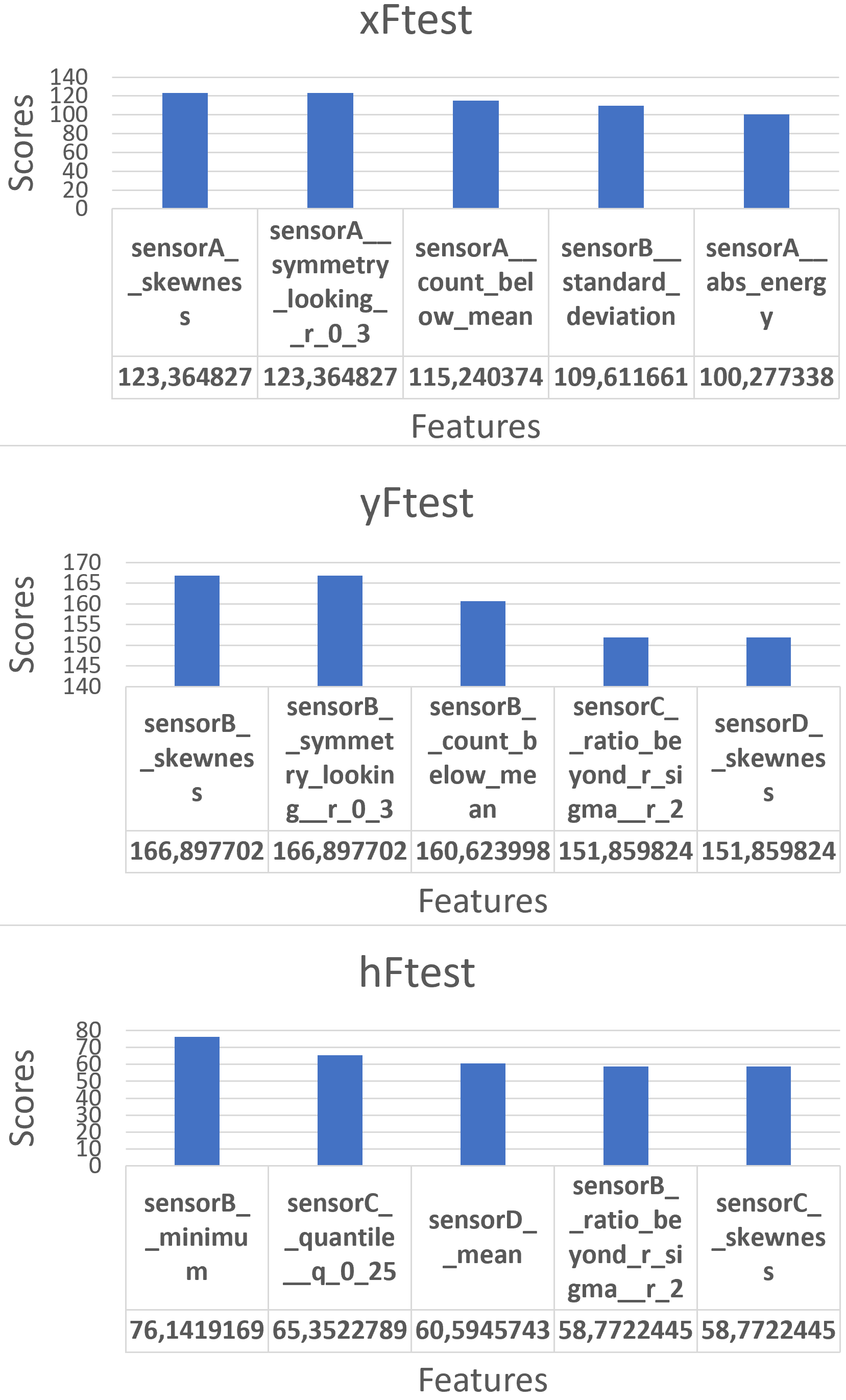}
    \caption{The five most important features for X-coordinate (xFest), for Y-coordinate (yFest), and for fall height (hFest).}
    \label{fig:feature-importance}
\end{figure}

Correlations between the extracted signal features and target columns is calculated by ANOVA F-test. The results of this test are used for feature selection, where the first five features that are strongly correlated to each of the target variable (x-coordinate, y-coordinate and fall height) are found. Each F-test evaluates the hypothesis that the response values grouped by predictor variable values are drawn from populations with the same mean against the alternative hypothesis that the population means are not all the same. A small p-value of the test statistic indicates that the corresponding predictor is important. The output scores is $-log(p)$. Therefore, a large score value indicates that the corresponding predictor is important. The most important features per X and Y-coordinates and fall height, along with their F-test values, are displayed in Figure \ref{fig:feature-importance}.




\section{Conclusions and Future Work}
In this work, we present a novel dataset, that concerns low-velocity, low-energy impact events in an acrylic, square thin plate and consists of  time series data from 4 piezoelectric transducers located at the corners of the plate. We share our dataset and the corresponding analysis to facilitate researchers in studying impact detection and localization in SHM applications (applying both supervised and unsupervised ML techniques), as our search for available experimental data did not lead us to any result that could serve our objectives. In the future, we plan to extend our dataset by conducting experiments with more types of impact sources and more setups of the same plate and sensors (as they are not identical and discrepancies may exist). We will also execute more experiments for a denser experimental space in order to compare previous results from models trained with less data and assess their potential to be improved. Finally, we opt to integrate IMU and strain sensors to cover additional and diverse SHM-related problems and requirements.

\footnotesize
\section*{Acknowledgements}
The present work was financially supported by the Foundation ``Andreas Mentzelopoulos''.

\normalsize
\bibliography{references}


\end{document}